# Pedestrian Path, Pose and Intention Prediction through Gaussian Process Dynamical Models and Pedestrian Activity Recognition

Raúl Quintero, Ignacio Parra, David F. Llorca, *Senior Member, IEEE*, Miguel A. Sotelo, *Fellow Member, IEEE*

*Abstract*—According to several reports published by worldwide organizations, thousands of pedestrians die in road accidents every year. Due to this fact, vehicular technologies have been evolving with the intent of reducing these fatalities. This evolution has not finished yet since, for instance, the predictions of pedestrian paths could improve the current Automatic Emergency Braking Systems (AEBS). For this reason, this paper proposes a method to predict future pedestrian paths, poses and intentions up to 1s in advance. This method is based on Balanced Gaussian Process Dynamical Models (B-GPDMs), which reduce the 3D time-related information extracted from keypoints or joints placed along pedestrian bodies into low-dimensional spaces. The B-GPDM is also capable of inferring future latent positions and reconstruct their associated observations. However, learning a generic model for all kind of pedestrian activities normally provides less accurate predictions. For this reason, the proposed method obtains multiple models of four types of activity, i.e. walking, stopping, starting and standing, and selects the most similar model to estimate future pedestrian states. This method detects starting activities 125ms after the gait initiation with an accuracy of 80% and recognizes stopping intentions 58.33ms before the event with an accuracy of 70%. Concerning the path prediction, the mean error for stopping activities at a Time-To-Event (TTE) of 1s is 238.01±206.93mm and, for starting actions, the mean error at a TTE of 0s is 331.93±254.73mm.

## I. Introduction

According to the *Annual Accident Report 2016* published by the European Road Safety Observatory, almost 26.000 people died in road traffic accidents in the European Union in 2014, including 5.729 pedestrians, which represent 22.09% of all fatalities. Concerning world statistics, data are more impressive. The *Global Status Report on Road Safety* published by the World Health Organization (WHO) in 2015 indicates that more than 1.2 million people died in road traffic accidents worldwide in 2013. About 275.000 of these fatalities were pedestrians.

Because of the high number of fatalities, during the last few years vehicles and infrastructures have been evolving to become intelligent machines with advanced technologies such as Assistive Intelligent Transportation Systems (AITS) [1], Pedestrian Protection Systems (PPS), AEBS or other sort of Advanced Driver-Assistant Systems (ADAS). Improving these technological advances is imperative because, as it is asserted effective avoidance maneuvers earlier. Despite this, not many he authors are with the Computer Engineering Department, Polytechnic School, University of Alcala, 28871, Alcala de Henares, Spain (e-mail: {raul.quintero, ignacio.parra, david.fernandezl, miguel.sotelo}@uah.es)

in [2], [3], the longer the braking initiation time, the higher the impact speed and thus, the injury risk. In fact, as claimed in the *Global Status Report on Road Safety* published by the WHO in 2015, an adult has less than a 20% chance of dying if struck by a car at less than 50 km/h but almost a 60% risk of dying if hit at 80 km/h. Hence, a precise assessment about the current and future pedestrian positions and an early detection of people entering a road lane is a major challenge in order to increase the effectiveness of AEBS. Similarly, an early recognition of pedestrian intentions can lead to much more accurate active interventions in the last second automatic maneuvers. In this way, in the last few years, with the aim of addressing these challenges, a lot of effort has been put into recognizing pedestrian activities and predicting trajectories and intentions.

This paper proposes a method to predict future pedestrian paths, poses and intentions up to 1s in advance using B-GPDMs, which reduce the 3D time-related positions and displacements extracted from keypoints or joints placed along the pedestrian bodies into low-dimensional latent spaces. The B-GPDM also has the peculiarity of inferring future latent positions and reconstructing the observation associated to a latent position from the latent space. However, as claimed in [4], learning a generic model for several kind of pedestrian activities normally provides less accurate predictions. To overcome this, the proposed method obtains multiple models of each type of pedestrian activity, i.e. walking, stopping, starting and standing, and selects the most appropriate to make predictions.

The remainder of this paper is organized as follows: Section II presents a brief survey of previous works focused on pedestrian intention and path prediction. Section III describes the dataset of pedestrian activities and the proposed system. The activity recognition and path prediction results obtained by the system are provided in Section IV. Finally, the conclusions and future research lines are discussed in Section V.

## II. Related Works

The problem of vision-based pedestrian detection for ADAS has been extensively researched in the past. As a consequence, many manufacturers have equipped their vehicles with commercial systems that warn the driver when a pedestrian or object is in front. Nonetheless, the estimation of future pedestrian states could improve these systems, allowing the activation of works that have been published so far about intention, path and pose prediction once pedestrians are detected.

To carry out these tasks, a wide range of features and information can be extracted from pedestrians. However, some of them are certainly more significant than others. In this sense, studies such as [5]–[13] give some useful clues. According to them, it seems that the most relevant features to compute path and intention predictions can be extracted from two sources. The first one corresponds to pedestrians since their motions, positioning information, orientations and head poses determine the variables that a driver commonly uses to infer intentions and to know whether pedestrians are aware of oncoming vehicles. The second source emerges from the situation criticality and the environment given that vehicle-pedestrian and curbside-pedestrian distances, existence of zebra crossing, or road width are significant data.

*A. Predictions using Pedestrian Features*

Regarding the first source of information, pedestrian motion features are regularly extracted to make predictions applying image processing instead of computing only pedestrian positions and velocities as several approaches do. E.g., in [14], [15], only positioning information is used to predict pedestrian-vehicle collisions and paths at short prediction horizons respectively. Additionally, in [16], [17], apart from using positioning information, augmented motion features derived from dense optical flow fields are also processed for path and intention predictions. The use of augmented motion features achieves more accurate results for stopping trajectories than using only positioning information. In these works, [16], [17], different approaches have been implemented and compared. An IMM-Kalman Filter (KF) is used to include two KFs for moving and non-moving pedestrians. Besides, a trajectory matching and filtering framework called Probabilistic Hierarchical Trajectory Matching (PHTM) is developed to compare the current pedestrian trajectory with trajectories previously learned. Apart from the two approaches described before, in [16] two GPDMs are also trained to make predictions, one for walking motions and the other for stopping activities. Both models are combined using an IMM-Particle Filter (PF) to select the most appropriate pedestrian dynamics. Another example of the use of motion features can be found in [18] where a method to recognise intentions from a moving vehicle is implemented using Support Vector Machine (SVM) classifiers. The motion features are gathered through the overlapping of pedestrian silhouette images at consecutive time steps.

Moreover, the orientations in which pedestrians are facing and head poses could be evaluated to predict future pedestrians positions. These features are investigated in [19], [20] in order to predict intentions. In [19], Histogram of Oriented Gradients (HOG) features are fed to an 8-class SVM classifier whose probabilities allow to model a Hidden Markov Model (HMM) to infer future orientations. In [20] it is presented an approach that combines intention recognition and path prediction by means of an IMM-EKF in combination with a Latent-Dynamic Conditional Random Field (LDCRF) to integrate positioning information and situational awareness computed by head pose estimation.

Furthermore, many dangerous situations arise when the driver's view of the road is obstructed by objects, making impossible the detection of pedestrians from the inside the vehicle. For this reason, infrastructure sensors in combination with roadside units can be mounted at urban hazard spots sending the appropriate signals to nearby vehicles through wireless communication channels. This solution is proposed in [21]–[23] with the aim of predicting crossing intentions using pedestrian motion features and linear 2-class SVM classifiers. Additionally, positioning information is extracted in [24]–[26] to create linear and non-linear velocity-time-based and position-based models which are able to predict paths in the course of a gait initiation at crosswalks or for typical pedestrian motions. Apart from using positioning information, heading angle is also considered in [27]. The work proposes a method based on the clustering of trajectories to avoid vehicle pedestrian collisions.

*B. Predictions using Context Information*

Despite urban environments are generally very complex, exploiting and analyzing the context information, i.e. the situational criticality and the spatial layout of the environment (structure of streets, sidewalks, intersections or crosswalks), can also provide some valuable information to AEBS. In this sense, this information and the pedestrian situational awareness are assessed in [28] by the vehicle-pedestrian distance at the expected collision point, the curbside-pedestrian distance and the pedestrian head orientation. The authors apply a Dynamic Bayesian Network (DBN) and a Switching Linear Dynamical System (SLDS) with the aim of predicting pedestrian paths from an approaching vehicle. Furthermore, pedestrian features and contextual information are also combined in [29]– [32]. The first work fuses two models to predict crossing intentions from a moving vehicle and computes contextual information such as lateral distances and times that pedestrians need to reach some goals (collision point, curbstone, ego lane or crosswalk) and pedestrian features such as positioning information, velocities and directions. In [30], curbside pedestrian and vehicle-pedestrian distances, head orientations, and pedestrian speeds are computed to predict intentions using a stereo-thermal camera mounted on the front-roof of a car. The work described in [31] is focused on identifying those features from the environment that are necessary to determine whether a pedestrian will cross the road at a crosswalk. Finally, a stereo-and infrastructure-based pedestrian detection system is presented in [32] to assess whether pedestrians will cross or wait to estimate their positions in a set of manually selected regions corresponding to the pedestrian waiting areas and the crossing region.

Likewise, as previously mentioned, many dangerous situations arise from the fact that the driver's view of the road scene may be obstructed by objects and, hence, it could be impossible to avoid a collision. Including prior knowledge about the scene such as objects, sidewalks, roads, entries and destinations might provide richer information to systems focused on predicting pedestrian trajectories. E.g., in [33], the task of inferring paths and intentions from a static camera is addressed by incorporating physical scene features and noisy tracker observations. Thereby, the effect of physical environments on pedestrian intentions is modelled through the information that is gleaned from physical scene features and prior knowledge of possible destinations.

*C. Discussion*

Although all these works obtain interesting results in path and intention prediction, applying pedestrian skeleton estimation offers a new approach to carry out these tasks, as recently proposed in [34]–[38]. Given that humans are not rigid objects, the motion analysis of each body part should be taken into account to make predictions since, e.g., whereas the motion of the head may not be relevant in starting intentions, a slightly motion of a knee could be indicative of that action.

Concerning modelling approaches, switching between models with different dynamics could be a successful option to achieve accurate predictions as proposed in several works. However, extensive experiments have not been carried out so far in order to fix the number of different pedestrian dynamics that could emerge in urban environments. Unlike other works that use two dynamical behaviors, in this paper a method based on four different pedestrian actions to obtain predictions is proposed. In relation to approaches which take into account past motion history to predict future paths, they may not be effective in situations where pedestrians suddenly appear in the vehicle trajectory. Therefore, although these systems obtain good predictions, they could not be useful in urban environments. The method proposed in this paper only needs two pedestrian observations to predict paths, poses and intentions.

Additionally, despite most works reviewed above are focused solely on predicting intentions, providing the probability of crossing with high confidence is not enough to avoid collisions. E.g., future pedestrian positions could be decisive in the computation of the best collision avoidance trajectory for an automatic steering system. In this paper pedestrian paths, poses and intentions are predicted with the aim of improving the AEBS.

III. SYSTEM DESCRIPTION

The proposed method is based on B-GPDMs, which reduce the 3D time-related positions and displacements extracted from keypoints or joints placed along the pedestrian bodies into low-dimensional latent spaces. The B-GPDM also has the peculiarity of inferring future latent positions and reconstructing the observation associated to a latent position from the latent space. Therefore, it is possible to reconstruct future observations from future latent positions. However, as claimed in [4], learning a generic model for all kind of pedestrian activities or combining some of them into a single model normally provides inaccurate estimations of future observations. For that reason, the proposed method learns multiple models of each type of pedestrian activity, i.e. walking, stopping, starting and standing, and selects the most appropriate one to estimate future pedestrian states at each time step. A general description of the method is shown in Fig. 1. A training dataset of motion sequences, in which pedestrians perform different activities, is split into 8 subsets based on typical crossing orientations and type of activity. Then, a B-GPDM is obtained for each sequence with one activity contained in the dataset. After that, in the online execution, given a new pedestrian observation, the current activity is determined using a HMM. Thus, the selection of the most appropriate model among the trained ones is centered solely on that activity. Finally, the selected model is used to predict the future latent positions and reconstruct the future pedestrian path and poses.

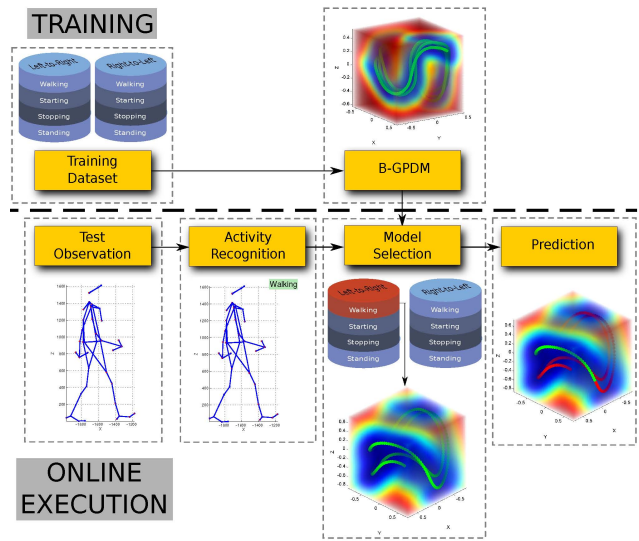

Fig. 1. General description of the method proposed. The algorithm is divided into two stages: offline training (top) and online execution (bottom).

*A. Dataset Description*

Two of the main goals of this work are to train accurate models with different pedestrian dynamics and to test the feasibility and limits of the proposed method in an extensive way under ideal conditions. For that purpose, a high frequency and low noise dataset published by Carnegie Mellon University (CMU) [39] has been used. On the one side, the high frequency of the dataset helps the algorithms to properly learn the dynamics of different activities and increases the probability of finding a similar test observation in the trained data without missing intermediate observations. On the other side, low noise models improve the prediction when working with noisy test samples.

The dataset contains sequences in which people are simulating typical pedestrian activities. 3D coordinates of 41 joints located along their bodies are gathered at 120 Hz. An example of a walking pedestrian observation from different points of view is shown in Fig. 2. Nevertheless, not all gathered joints offer discriminative information about the current and future pedestrian activities. In fact, joints located along the arms do not contribute to distinguish walking, starting, stopping or standing activities. For that reason, a subset of 11 joints was selected in order to infer future pedestrian states. An example of a pedestrian observation of this subset from different points of view is shown in red markers in Fig. 2.

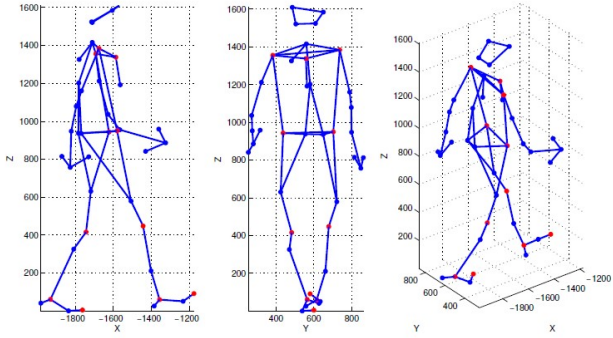

Fig. 2. Pedestrian observation extracted from the dataset published by CMU in which 41 joints, represented in blue markers, are shown. The subset composed of 11 joints is represented in red markers.

The CMU dataset contains a considerable number of different activities, including human interaction of several subjects. We pre-select and filter the available sequences according to the following criterion: only sequences including walking, starting, stopping and standing activities without orientation changes of the subjects are selected. This way, a total of 490 sequences composed of 302,470 pedestrian poses from 31 different subjects were extracted. Hereafter, this set of sequences will be named as CMU-UAH dataset.

After this extraction, the CMU-UAH dataset was divided into 8 subsets. The first division was based on the orientation of typical crossing activities, i.e. left-to-right and right-to-left. The second one was based on the type of activity, i.e. walking, starting, stopping and standing. Those sequences with more than one activity were cropped into subsequences with only one action. However, none of the works reviewed on Section II offers a discussion on how to identify the instant that a pedestrian starts walking or finishes an activity of crossing, starting or stopping. Consequently, a guideline will be proposed in this paper later on. A breakdown of the CMUUAH dataset based on the number of sequences and pedestrian observations is shown in Table I.

TABLE I  Number of sequences and number of observations for each type of activity.

| | Sequences | | | | |
|---|---|---|---|---|---|
| Orientation | Walking | Starting | Stopping | Standing | Total |
| Left-right | 240 | 142 | 56 | 224 | 662 |
| Right-left | 191 | 121 | 27 | 156 | 495 |
| Total | 431 | 263 | 83 | 380 | 1157 |
| | Observations | | | | |
| Orientation | Walking | Starting | Stopping | Standing | Total |
| Left-right | 107324 | 10732 | 2522 | 43151 | 163729 |
| Right-left | 95113 | 10940 | 1276 | 31412 | 138741 |
| Total | 202437 | 21672 | 3798 | 74563 | 302470 |

It is worth remarking that each pedestrian observation is composed of pose and displacements. The former is concerned with the 3D position of each joint and the latter are associated with the displacement of each joint between two consecutive iterations. In practice, the joint displacements are key features since they increase the feasibility of reconstructing future pedestrian paths and improve the accuracy of the pedestrian activity classification.

*1) Event-labelling Methodology:* The guideline of event labelling proposed in this paper allows to identify the instant that a pedestrian starts or finishes an activity. Specifically, a starting activity is defined as the action that begins when the pedestrian moves one knee to initiate the gait and ends when the foot of that leg touches the ground again. A walking activity is defined as the action that happens after a starting activity and before a stopping activity. Moreover, a stopping activity is defined as the action that begins when a foot is raised for the last step and finishes when that foot treads the ground. Finally, standing activities are defined as the actions that happen after stopping activities and before starting activities. These criteria were adopted because these transitions are easily labelled by human experts, thus enabling the creation of reliable ground-truths.

### B. Pedestrian Skeleton Estimation

A pedestrian skeleton estimation algorithm based on point clouds extracted from a stereo pair and geometrical constraints was implemented to test the proposed method with noisy observations. This algorithm is a variation of the method proposed in [35], [40] and it is described in [38]. An example of a pedestrian skeleton estimation is shown in Fig. 3.

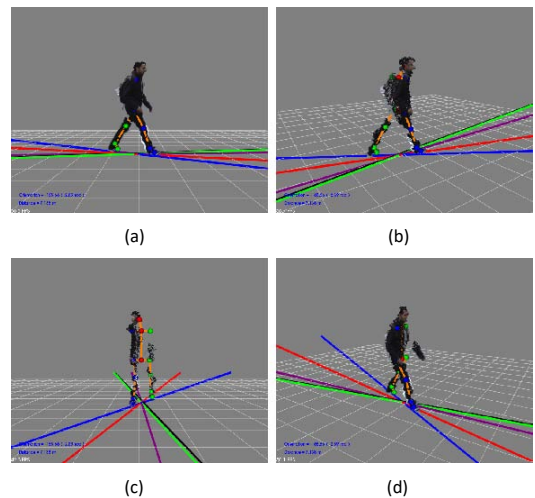

Fig. 3. Example of a pedestrian skeleton estimation. Green markers correspond to left joints, blue markers to right joints and red markers to head, center of shoulders and center of hips. The lines represent the pedestrian heading computed from each body part.

### C. Learning Pedestrian Activities

As mentioned above, this paper describes a method based on the B-GPDM, which is a modified version of the Gaussian Process Dynamical Model (GPDM), to learn 3D time-related information extracted from pedestrian joints in order to predict paths, poses and intentions. The GPDM, described in [4], [41], provides a framework for transforming a sequence of feature vectors, which are related in time, into a low dimensional latent space. In order to apply this transformation, the observation and dynamics mappings are computed separately in a non-linear form, marginalizing out both mappings and optimizing the latent variables and the hyperparameters of the kernels. The incorporation of dynamics not only allows to make predictions about future data but also helps to regularize the latent space for modelling temporal data. Therefore, if the dynamical process defined by the latent trajectories in the latent space is smooth, the models will tend to make good predictions. Likewise, given a latent position from the latent space, the associated observation can be reconstructed. Nonetheless, learning a generic model for all kind of pedestrian activities or combining some of them into a single model could produce poor dynamical predictions as claimed in [4]. For that reason, the proposed method learns multiple models for each type of pedestrian activity, i.e. walking, stopping, starting and standing, and selects the most appropriate among them to predict future pedestrian states at each time step. The learning stage starts loading all sequences contained in the CMU-UAH dataset. Because of the coordinate system of these sequences is referenced to the sensor, the 3D translation parameters of each observation are removed so that the origin of the reference system is relocated in the pedestrian. This allows the algorithms to deal with pedestrians regardless of their positions with respect to the sensors. After that, the variables are scaled by subtracting the mean and dividing each one by its standard deviation in order to have zero-mean and unit-variance data.

Since the B-GPDM requires iterative procedures for minimizing the log-posterior function, the latent positions X, the hyperparameters θ and β, and the constant κ have to be properly initialized. On the one hand, the latent coordinates are initialized by PCA and, on the other hand, the kernel parameters and κ are initialized by using the values proposed in [4]. Finally, a B-GPDM is learned for each sequence. An example of a model corresponding to a pedestrian that is walking 6 steps is shown in Fig. 4. The green markers indicate the projection of the pedestrian observations onto the subspace. The model variance is represented from cold to warm colors. Whereas a high variance (warm colors) indicates that illogical pedestrian observations can be reconstructed, a low variance (cold colors) indicates that pedestrian observations similar to the learned sequence may be obtained from a latent position or to the end of a starting sequence. Thus, if the previous activity were recognized as walking, then the next dynamics would be determined as walking or stopping, not as starting. Thereby, the process of how a pedestrian changes its dynamics over time can be described by a Markov Process. At any time, the pedestrian can do one of a set of 4 distinct actions (s). These activities or states are not observable since only 3D information from joints belonging to the pedestrian is available. Therefore, the states can be only inferred through the observations (x). For this reason, the implementation of a first-order HMM allows to model the transitions between activities and to recognise the correct one taking into account the previous dynamics. The Viterbi algorithm is a dynamic programming procedure for finding the most likely state sequence given an observation sequence. This way, choosing sequences of a single element, the probability of an observation x of being in the j-th state of s at an instant of time $t$ is formulated as:

$$p(\mathbf{s}_j^t|\mathbf{x}^t) = \frac{p(\mathbf{x}^t|\mathbf{s}_j^t)p(\mathbf{s}_j^t)}{\sum_{i=1}^{4} p(\mathbf{x}^t|\mathbf{s}_i^t)p(\mathbf{s}_i^t)} \quad (1)$$

where $p(s^t_i)$ represents the prior probability and $p(x^t|s^t_j)$ the emission probability. The prior probability is computed as:

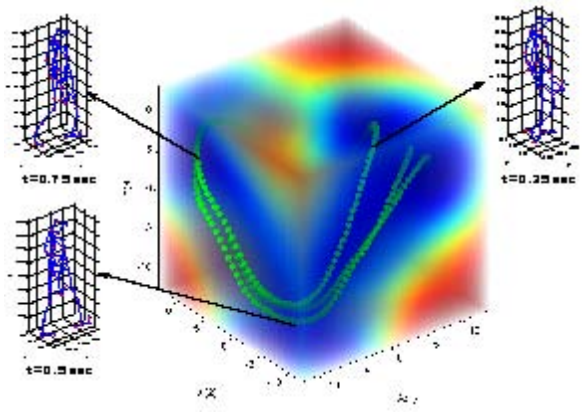

Fig. 4. B-GPDM corresponding to a pedestrian that walks 6 steps. The projection of the pedestrian motion sequence onto the subspace is represented by green markers. The model variance is indicated from cold to warm colors.

### D. Activity Recognition

Since several models with different dynamics are previously trained, the activity recognition for the current pedestrian observation allows to select afterwards the most accurate mode to estimate future pedestrian states. Naive-Bayes classifiers as proposed in [37], or the maximum similarity between the current observation and each observation of the training dataset may determine the activity. Nevertheless, in the last case, if the maximum similarity was applied directly, i.e., without modelling the evolution of the pedestrian activity, higher errors would be achieved when selecting the most appropriate model due to the likeness between observations of different dynamics. E.g., an observation of a pedestrian that is walking may be similar to an observation belonging to the beginning of a stopping sequence

$$p(\mathbf{s}_j^t) \propto \max_{i=1}^{4}[p(\mathbf{s}_j^t|\mathbf{s}_i^{t-1})p(\mathbf{s}_i^{t-1}|\mathbf{x}^{t-1})], \quad t > 1 \quad (2)$$

where $p(s^t_j | s^{t-1}_i)$ corresponds to the probability of changing from the i-th to the j-th state defined by means of a Transition Probability Matrix (TPM). The values of transitions between states were experimentally fixed maximising the success rate. $p(s^{t-1}_i | x_{t-1})$ corresponds to the probability of being in the i-th state of s at the previous instant. The initial probability $p(s^t)$ is uniformly distributed since the pedestrian activity is unknown in t = 1. The emission probability $p(x_t|s^t_j)$ is defined as:

$$p(\mathbf{x}^t|\mathbf{s}_j^t) \propto \max_{i=1}^{N}\left(\frac{1}{1+\alpha_i} + \frac{1}{1+\beta_i}\right) \quad (3)$$

where $\alpha_i \in [0, \infty]$ and $\beta_i \in [0, \infty]$ correspond to the Sum of Squares Errors (SSE) for the pedestrian pose and the joint displacements respectively. The SSE are computed between the current pedestrian observation x and the N observations of the training data subset belonging to the j-th state of s. Before computing $\alpha_i$, the pose of the current pedestrian observation and the poses of the training observations are scaled and referenced to the same joint. The scale factor applied to each observation is obtained by the sum of ankle-knee and knee-hip distances. The displacements are not scaled with the intent of finding pedestrians with similar joint velocities.

*E. Path, Pose and Intention Prediction*

Once the pedestrian activity in t has been estimated, the selection of the most appropriate model allows to make accurate predictions about paths, poses and intentions. For these tasks, a search of the most similar training observation and its model is computed. This observation corresponds to the i-th element for the activity s in Equation 3. After that, the latent position that represents the most similar observation is used as the starting point for a more accurate search in the selected model applying a gradient descent algorithm. Due to the fact that close points in the latent space are also close in the data space, it is expected that a more similar non-trained observation can be found around this starting point. The function that is minimized in the gradient descent algorithm is defined as:

$$\epsilon(\mathbf{x}) = \sum_{j=1}^{d}((\mathbf{y} - \boldsymbol{\mu})^2) + \frac{1}{2}\sum_{j=1}^{q}(\mathbf{x}^2) \quad (4)$$

where y is the current pedestrian observation and μ represents the pedestrian observation reconstructed from the latent position x. Both observations are previously scaled and referenced to the same joint. Finally, d corresponds to the dimension of the original observation and q to the dimension of the model.

Once the final latent position has been estimated, predictions of N latent coordinates are iteratively made and their associated observations are reconstructed. Thereby, given the current pedestrian location with respect to the sensor, the future pedestrian path can be computed adding the consecutive N predicted displacements. It is noteworthy that the reference point to reconstruct the path is the right hip since it corresponds to a point close to the center of gravity. Additionally, given the N future pedestrian observations, the future intentions can be estimated through the application of the activity recognition algorithm, explained in Section III-D, to these observations.

IV. RESULTS

Throughout this section, the main results of the algorithms described above are discussed. All algorithms were tested using the CMU-UAH dataset adopting a one vs. all strategy. This means that all the models generated by one test subject were removed from the training data before performing tests on this subject. Because of the pedestrian displacements are computed from two consecutive poses, 301.980 observations are finally analyzed. Additionally, the activity recognition and prediction algorithms were also tested using a sequence of pedestrian data extracted by the skeleton estimation algorithm.

*A. Activity Recognition Results*

The activity recognition results are summarized on a confusion matrix shown in Table II. Nonetheless, a more exhaustive data assessment is represented in Table III where the different activity recognition performances are compared taking into account the pedestrian features, number of joints and activity.

TABLE II CONFUSION MATRIX USING 11 PEDESTRIAN JOINTS.

| | | Predicted | | | |
| --- | --- | --- | --- | --- | --- |
| | | Standing | Starting | Stopping | Walking |
| Actual | Standing | 72011 | 1396 | 174 | 68 |
| | Starting | 1451 | 13313 | 13 | 68 |
| | Stopping | 126 | 0 | 1951 | 17 |
| | Walking | 262 | 494 | 1508 | 200 |

TABLE III EVALUATION OF ACTIVITY RECOGNITION PERFORMANCE WITH RESPECT TO PEDESTRIAN FEATURES, NUMBER OF JOINTS AND ACTIVITY.

| Features | | Pose + Disp | | Pose | | Disp | |
| --- | --- | --- | --- | --- | --- | --- | --- |
| Joints | | 41 | 11 | 41 | 11 | 41 | 11 |
| Accuracy | | 90.69% | **95.13%** | 88.39% | 91.28% | 94.76% | 94.23% |
| Precision | Standing | 89.77% | 97.51% | 88.93% | 95.54% | 97.27% | **98.04%** |
| | Starting | 77.88% | **87.57%** | 66.30% | 79.38% | 82.79% | 83.96% |
| | Stopping | 44.59% | **53.51%** | 41.78% | 40.06% | 35.79% | 35.72% |
| | Walking | 92.50% | **95.57%** | 89.89% | 91.35% | 94.90% | 94.81% |
| Recall | Standing | 88.85% | 96.97% | 84.31% | 87.86% | 97.01% | **97.19%** |
| | Starting | 48.60% | **61.49%** | 31.33% | 39.14% | 52.62% | 54.90% |
| | Stopping | 36.45% | **51.38%** | 32.87% | 37.11% | 41.14% | 40.90% |
| | Walking | 96.90% | 98.88% | 97.04% | **99.13%** | 98.43% | 98.36% |
| F1-Score | Standing | 89.31% | 97.24% | 86.56% | 91.54% | 97.14% | **97.61%** |
| | Starting | 59.85% | **72.25%** | 42.55% | 52.43% | 64.34% | 66.39% |
| | Stopping | 40.11% | **52.42%** | 36.79% | 38.53% | 38.28% | 38.13% |
| | Walking | 94.65% | **97.20%** | 93.33% | 95.08% | 96.63% | 96.55% |

*1) Joints Influence on the Performance:* The results verify that shoulder and leg motions, which are associated with the 11 joints, are more valuable sources of information than other body parts to recognise the current pedestrian action. E.g., including the arms and upper body parts do not improve the results, probably because they do not introduce distinctive information about them. Specifically, the maximum accuracy, 95.13%, is achieved when the observations are composed of poses and displacements from only 11 joints. However, the accuracy falls to 90.69% when 41 joints are used. Considering only body poses, a similar conclusion is drawn since the maximum accuracy is 91.28% and 88.39% for 11 and 41 joints, respectively. Finally, when the observations are composed solely of pedestrian displacements, the activity recognition results are not significantly influenced by the number of joints.

*2) Features Influence on the Performance:* Regarding the distinction among activities, the displacements perform a better recognition of standing actions from the rest of activities. However, with respect to starting and stopping actions, a higher number of critical missclassifications are produced. This means that the displacements do not allow to reliably distinguish whether a pedestrian is carrying out the first or last step. The poses and displacements offer a more discriminative information in these cases.

Considering the body pose as the only feature, standing actions are repeatedly recognized as walking activities since, when the pedestrian legs are closed, the poses from both states are very similar in those instants of time. Therefore, the displacements are valuable information in these cases.

Regarding the observations composed of body poses and displacements, the most frequent missclassifications are produced by delays or pedestrians with low-speed motions. The first cause is related to the event-labelling methodology selected by the human expert. It seems that the first half of the first step and the second half of the last step contain the most perceptible information to determine starting and stopping actions respectively. Hence, the rest of these steps is normally recognized as walking action. Concerning the second cause, walking activities are recognized as starting or stopping actions when pedestrians with low-speed motions are tested. All these last missclassifications are not critical from the point of view of the path estimation since these actions have similar dynamics. Likewise, the beginning of a starting action and the ending of a stopping motion contains body poses which are equivalent to poses labelled as standing actions. Hence, a significant number of missclassifications are also produced between these activities.

11 joints along with the ground-truth are illustrated. Several examples of pedestrian poses at different instants of time are illustrated at the top of the figure. These poses are represented in different colours according to the classification result. Black represents standing, green starting, red walking and blue stopping. In the middle, the probabilities of each activity at each instant of time are shown. Finally, at the bottom, a zoom in of the transitions is illustrated. The graph shows short delays in the standing-starting and stopping-standing transitions. These delays will be discussed later. On the other hand, throughout walking actions, local maxima and local minima of walking probabilities appear in the graph when the pedestrian legs are open and closed, respectively.

*3) Labelling Influence on Delays:* In Table IV and Fig. 6, the transitions from activities are analyzed in detail. This analysis is focused on the accuracy of transition detection and its delays. The evaluation criteria fix a range of [−500, 500] ms around the event labelled by the human expert. Within this range, a multiframe validation algorithm is applied in order to ensure the transition detection and reduce false positive changes produced by missclassifications. The number of frames is fixed to 6 (50ms). Thereby, the algorithm detects a transition when 6 consecutive pedestrian observations are recognized as the same activity but this is different to the action classified in t − 6. Finally, the activity detection delay is computed from the instant of time where the event was marked by the human expert and the instant of time where the transition was detected by the algorithm.

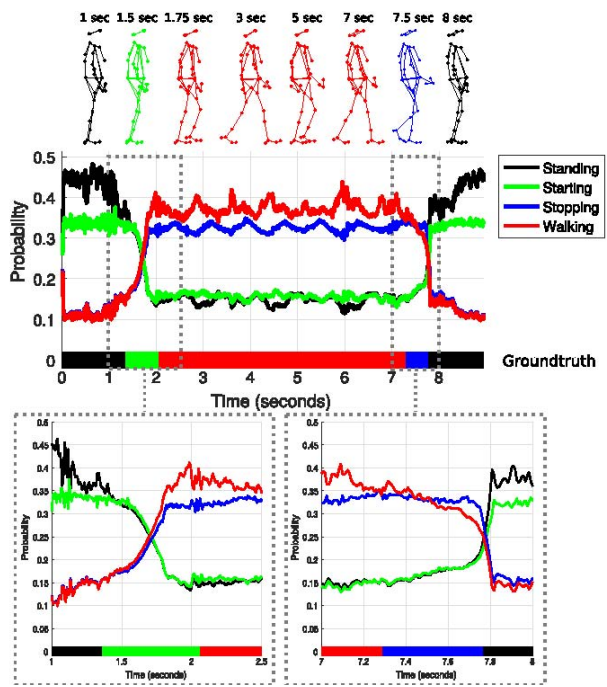

Fig. 5. Example of activity recognition using poses and displacements extracted from 11 joints. Top: pedestrian poses at significant instants of time. Middle: probabilities for each activity. Bottom: zoom in of the transitions.

A graphical example of several of the previous statements is shown in Fig. 5 where the classification probabilities using

TABLE IV Accuracy of detected and non-detected transitions.

| Transition | Accuracy | |
| --- | --- | --- |
|  | 41 Joints | 11 Joints |
| **Standing - Starting** | 71.60% | 97.94% |
| **Starting - Walking** | 83.97% | 95.42% |
| **Walking - Stopping** | 62.20% | 74.39% |
| **Stopping - Standing** | 78.85% | 91.25% |
| **Overall** | 76.16% | 93.25% |

When 11 joints are used, the number of transitions correctly and incorrectly detected is 622 and 45 respectively, i.e. the accuracy is 93.25%. Most of the transitions which are not detected corresponds to walking-stopping changes. This occurs due to the fact that the number of observations in the dataset belonging to a stopping activity is significantly smaller than other actions and stopping steps are usually faster than starting steps. The mean lengths of time of starting and stopping steps along with their standard deviations are 686.06 ± 202.91 and 381.22 ± 78.92 ms respectively.

Regarding the delays of the detected transitions, the results show that these are not significantly influenced by the number of joints since the multiframe validation algorithm filters most of the missclassifications. It should also be pointed out that starting-walking transitions have negative delays since the first half of the first step contains the most perceptible information to determine starting actions.

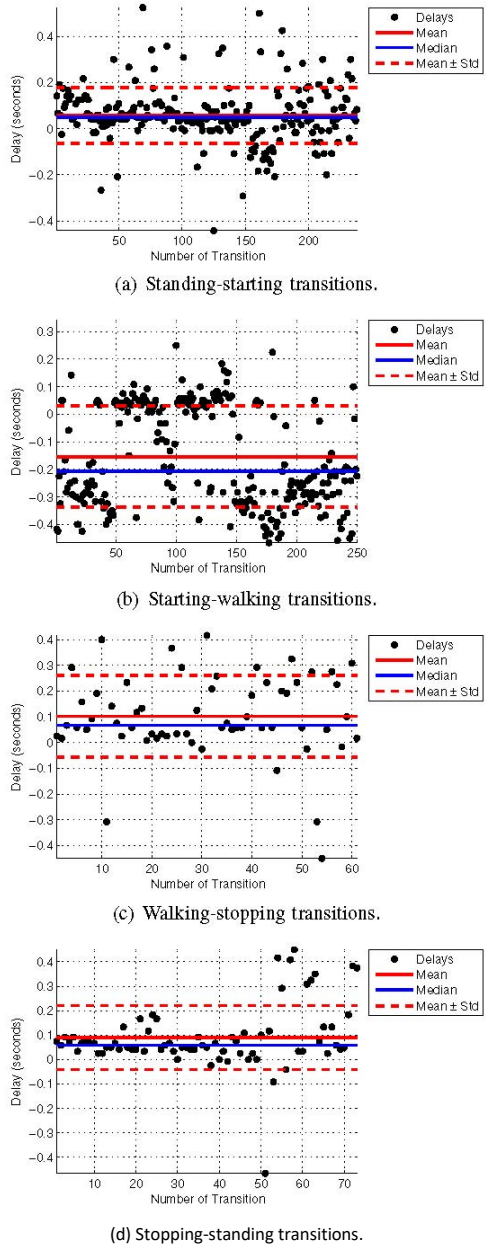

(a) Standing-starting transitions.

(b) Starting-walking transitions.

(c) Walking-stopping transitions.

(d) Stopping-standing transitions.

Fig. 6. Delays in seconds of detected transitions when 11 joints are used. The graphs show the delays of each transition along with the mean, median and standard deviation values.

The method proposed in this document recognizes starting intentions 125ms after the gait initiation with an accuracy of 80% when 11 joints are considered. These results are similar to the delays achieved in [18], [21]. Nonetheless, a multiframe validation of 50ms is carried out in order to filter missclassifications and a higher number of different dynamics are modelled in the proposed method. This means that the consideration of only one transition, i.e. standing-walking, instead of two dynamical changes, i.e. standing-starting and starting-walking, could accomplish better results. However, if only two states were considered the path prediction could be negatively influenced.

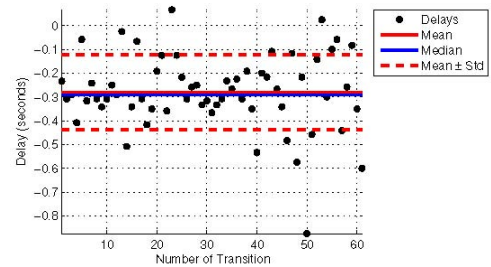

Fig. 7. Delays from walking-stopping transitions to standing events.

Additionally, an analysis of delays from walking-stopping transitions to the standing events labelled by the human expert was done and showed in Fig. 7. This analysis is important in order to know the delay from a stopping detection until the real standing event. The analysis shows that a mean delay of −279.92 ± 158.59ms is achieved in the detection of walking-stopping transition before a standing event. In fact, most standing events can be predicted a few tens of ms in advance. Specifically, the method proposed in this document recognizes stopping intentions 58.33ms before the event with an accuracy of 70% when 11 joints are considered. This data is slightly worse than the results accomplished in [16]–[18] due to the non-detection of walking-stopping transitions previously discussed.

### B. Activity Recognition using Vision-based Skeletons

The activity recognition was also examined using a sequence of noisy observations extracted by the pedestrian skeleton estimation algorithm. In Fig. 8, it is represented images extracted from the sequence that corresponds to a pedestrian that stops walking and starts walking again on a zebra crossing from the right to left.

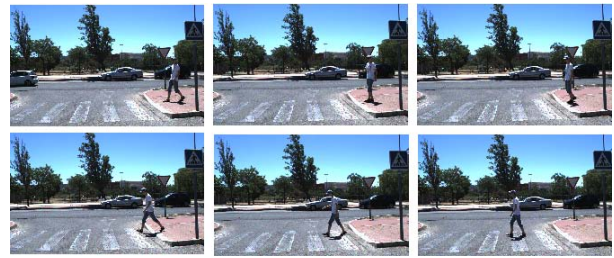

Fig. 8. Frames of a sequence in which a pedestrian crosses a road.

In Fig. 9, the activity recognition is represented. At the top of the figure, the pedestrian point clouds extracted by the pedestrian segmentation algorithm and the skeleton estimation at different instants of time are shown. The graph shows that the standing action at the curb was correctly recognized. However, the walking-stopping was not detected due to the usage of noisy observations.

### C. Pedestrian Path Prediction Results

Throughout this section, the evaluation of path prediction results is performed considering 11 joints. Firstly, the outcomes of this task using the CMU-UAH dataset are shown. After that, the path prediction using noisy observations extracted by the skeleton estimation algorithm is analyzed.

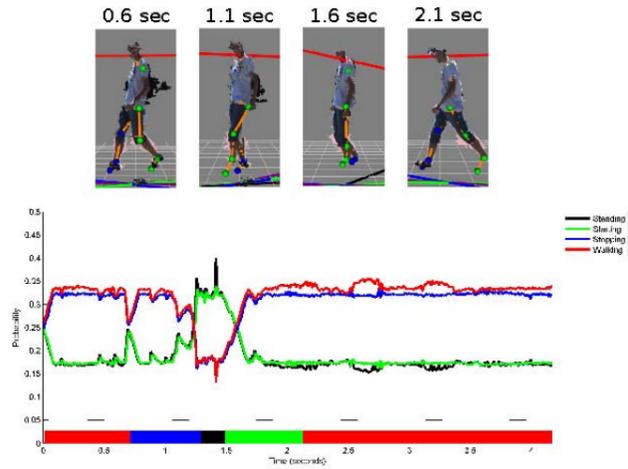

Fig. 9. Activity recognition when poses and displacements extracted from the skeleton estimation algorithm are used. Top: pedestrian poses at significant instants of time. Bottom: probabilities for each activity.

*1) Pedestrian Path Prediction Results:* As explained in Section III-E, once the pedestrian activity is estimated, the most appropriate model is selected and the prediction of future observations is iteratively performed using that model. Accordingly, a good path prediction depends strongly on a good activity recognition. The path prediction evaluation is performed using the activity recognition results discussed in Section IV-A. In this evaluation, the Mean Euclidean Distance (MED) between the predicted pedestrian locations and the ground-truth for time horizon values up to 1s are analysed. Due to the fact that the most dangerous traffic situations usually happen when the pedestrians start to cross or when they stop before crossing, the evaluation is done around these situations. Thereby, the MED are computed at different TTE, i.e. time to start walking and time to stop walking. Positive TTE values make reference to instants of time before the event and negative values to instants of time after the event.

In Table V and Fig. 10, the combined longitudinal and lateral MED along with the standard deviation are shown. Regarding starting activities, the errors before the event are mainly produced due to the fact the algorithm assumes zero displacements when the pedestrian activity is recognized as standing, however, this is not the case in the ground-truth since small motions were gathered. The errors after the event exponentially grows up since the recognition of a starting activity has a mean delay around 60ms and the pedestrian is accelerating. However, when the pedestrian finishes to speed up, the MED tend to be linear. Additionally, due to the fact that the B-GPDM is a dimensionality reduction technique, the errors are not significantly influenced by the number of joints. In order to contextualize the errors, the mean displacement for starting activities belonging to the CMU-UAH dataset was computed. Throughout a starting activity, the pedestrian has a mean displacement value of 193.98±78.52mm. Likewise, the mean displacement at 1s after and before the event is 467.92±264.97 and 41.24±67.91 mm respectively. It is worth mentioning that other dynamical changes could happen within the TTE range of [1-0] ss. E.g., a stopping-standing transition could be carried out by the pedestrian a few hundreds of ms before the event.

TABLE V Combined longitudinal and lateral MED±STD in mmat different TTE for predictions up to 1s using 11 joints.

| TTE (sec) \ Prediction (sec) | Standing-Starting | | Stopping-Standing | |
|---|---|---|---|---|
| | 0.5 | 1 | 0.5 | 1 |
| 1 | 15.33 ±17.55 | 38.86 ±54.07 | 90.94 ±103.91 | 238.01 ±206.93 |
| 0.5 | 28.33 ±33.52 | 141.79 ±140.89 | 150.83 ±223.89 | 462.06 ±567.53 |
| 0 | 89.10 ±88.38 | 331.93 ±254.73 | 100.66 ±88.64 | 244.23 ±250.99 |
| -0.5 | 116.00 ±113.39 | 296.23 ±228.83 | 20.16 ±19.49 | 64.34 ±95.74 |
| -1 | 79.17 ±93.07 | 161.14 ±186.36 | 51.24 ±67.85 | 183.66 ±183.17 |

The results focused on starting activities are similar to the results achieved in other works reviewed in Section II. Specifically, in [25] a MED value of 315mm is accomplished for a time horizon of 1.2 ss. This value is similar to the value obtained by the approach described in this paper for a TTE of 0s and a time horizon of 1s (331.93mm). Nonetheless, the event-labelling and prediction evaluation methodologies proposed in that work changes with respect to the described in this document. In [24], the MED at a starting event for a time horizon of 0.6s is 80mm. The method described in this paper achieves a MED value of 89.1mm at the instant of a starting event for a time horizon of 0.5 ss. In [26], a RMSE value of 334mm at 1s is obtained, this value is slightly lower than the RMSE obtained by the approach described in this document for a TTE of 0s and a time horizon of 1s (418.09mm). However, the predictions of this work are evaluated for all time steps instead of being assessed at different TTE and need a temporal window of n trajectory points to be performed instead of using two observations as the method described in this paper does.

Regarding stopping activities, the errors before the event tend to be linear since the mean length of stopping steps are 381.22±78.92 ms and the second half of the last step contain the most perceptible information to determine stopping actions. Thereby, an appropriate model could not be chosen until a few hundreds of ms before the event. After the event, the error decreases and tends to be logarithmic. However, at a TTE value of -1s, the errors grow up due to the fact that a new pedestrian dynamical change could happen. Once again, in order to contextualize the errors, the mean displacement for stopping activities belonging to the CMU-UAH dataset was computed. Throughout these activities, the pedestrian has a mean displace-ment value of 164.37±63.33mm. Likewise, the mean displacement at 1s after and before the event is 102.15±63.50 and 679.15.37±306.77mm respectively.

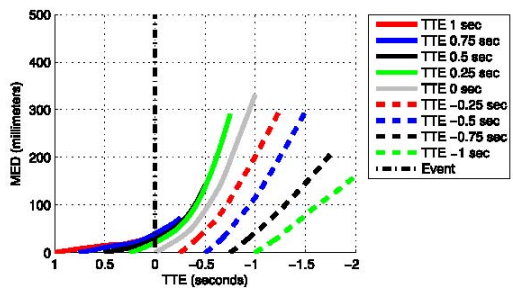

(a) For starting events and 11 joints

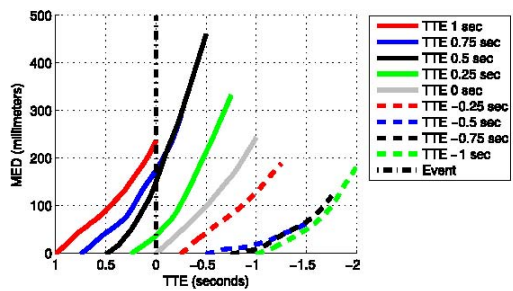

(b) For stopping events and 11 joints

Fig. 10. Combined longitudinal and lateral MEDs in mm at different TTEs$_{sr}$ for predictions up to 1s.

Comparing the results with the outcomes achieved by other works, these are similar. In particular, in [25], a MED value of 224mm is accomplished for stopping activities at 1.2 ss. The method proposed in this paper achieves a MED value of 238.01mm for a TTE of 1s and a time horizon of 1s. In [26], a RMSE value of 292mm at 1s is obtained, this value is slightly lower than the RMSE obtained by the approach described in this document for a TTE of 1s and a time horizon of 1s (314.5mm). However, the algorithm described in that work needs a temporal window of n trajectory points to perform the predictions instead of using two observations as the method described in this paper does. Moreover, the predictions are evaluated for all time steps instead of being assessed at different TTE. In [20], the lateral MED for a time horizon of 1s at 1s before the event is 140±180mm. The method described in this document achieves a lateral MED value of 226.99±208.01mm.

*D. Path Prediction using Vision-based Skeleton Estimation*

In this section, the path prediction is examined using the sequence of noisy observations described in [38]. In Fig. 11, the MED in mm for predictions up to 1s when poses and displacements computed from the sequence by using the skeleton estimation algorithm are represented. The method achieves lateral MED values of 131.71±57.89, 250.95±89.00, 355.80±123.37 and 448.84±157.39mm at 0.25, 0.5, 0.75 and 1s respectively. However, larger combined lateral and longitudinal MED are obtained. This is due to the fact that the pedestrian is not walking perpendicular to the sensor. As explained in Section III-A, the training dataset is composed of people with left-to-right and right-to-left heading with a variance in the longitudinal component close to zero. Hence, the future path reconstruction is corrupted by the predicted displacement vectors. To solve this problem, the observations in the training set and test set should be normalized by means of rotations to have the same orientation with respect to the sensor. In this way, the method could predict future paths regardless of the pedestrian direction.

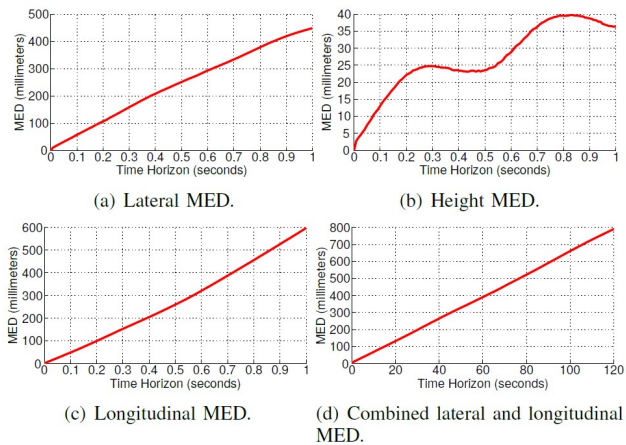

Fig. 11. MED in mm for predictions up to 1s using a sequence of noisy observations.

*E. Processing Time*

This section resumes the processing times of each step carried out by the method. The training stage has been performed using MATLAB 2014 64-bits with a processor Intel i7-2600K 3.40GHz. As mentioned in [4], the computational bottleneck for the B-GPDM is the inversion of the kernel matrices, which is necessary to evaluate the likelihood function and its gradient. As expected, the longer the sequence, the higher the processing time due to the fact that the dimensions of the kernel matrices depends on the number of samples in the sequence. For this reason, the processing time tends to be exponential with the number of samples in the sequences.

TABLE VI
PROCESSING TIMES IN MS OF EACH STEP PER PEDESTRIAN OBSERVATION.

| | Joints | 41 | 11 |
|---|---|---|---|
| Activity Recognition | Mean | 85.0 | 43.6 |
| | Std | 29.7 | 21.0 |
| Path Prediction | Mean | 868.3 | 829.7 |
| | Std | 1284.4 | 1232.6 |
| Total | Mean | 741.5 | 670.5 |
| | Std | 1183.2 | 1129.8 |

The path prediction and activity recognition has been performed by means of MATLAB 2016 64-bits with a processor Intel i7-7700K 4.20GHz. The processing times are showed in Table VI.

The path prediction depends on the model selected in order to estimate the future pedestrian trajectory. If this model corresponds to a long sequence, the processing time is higher because the path prediction compute the inversion of the kernel matrix, which is necessary to evaluate the likelihood function and its gradient between the test observation and the reconstructed observation from the model. The mean total processing time is shorter than the mean path prediction time due to the fact that when the activity is recognized as standing, the path prediction is not performed.

## V. Contributions and Future Work

This paper proposes a method to predict future pedestrian paths, poses and intentions up to 1s in advance using B-GPDMs. Additionally, an exhaustive assessment about activity recognition and path prediction algorithms has been performed. Concerning activity recognition, the results verify that shoulder and leg motions are more valuable sources of information than other body parts to recognise the current pedestrian action. Specifically, the maximum accuracy, 95.13%, is achieved when observations composed of a few joints placed along these body parts are taken into consideration.

Moreover, at least two types of features are needed in the action recognition when more than two dynamical behaviors are considered, i.e. body poses and displacements. Regarding this task, the method proposed in this document detects starting intentions 125ms after the gait initiation with an accuracy of 80% and recognizes stopping intentions 58.33ms before the event with an accuracy of 70% when joints from shoulders and legs are considered.

Concerning the path prediction results, similar errors are obtained with respect to other works. The measure of accuracy chosen for the path evaluation is the MED at different TTE that gives objective information of the path prediction performance. Although other works accomplished slightly smaller errors than the method proposed in this document, their prediction algorithms need a temporal window of n trajectory points instead of using two observations and the errors are evaluated for all time steps instead of being assessed at different TTE.

From the results and conclusions of the present work, several lines of work can be proposed. E.g., a higher number of sequences should be considered including in the dataset children or elderly people. It is also necessary testing all algorithms with different type of features or combining them. Since the performance of the stopping activity is not satisfactory enough due to the low number of samples, some method for imbalance data can be adopted to solve this problem. In a higher level, the combination of context-based information along with a situation criticality evaluation and a pedestrian body language analysis would allow to develop more reliable AEBS. Thus, scene understanding, pedestrian detection and prediction algorithms are interesting lines of research in the ITS field. In order to obtain more accurate pedestrian skeletons, markerless motion capture approaches or pose estimation based on CNN such as the algorithms proposed in [42], [43] could be developed instead of algorithms based on geometrical constrains. On the other hand, creating a extensive dataset of real pedestrian situations would make possible to compare different approaches in similar conditions. The event-labelling methodology proposed in this paper would help to human experts determine the different pedestrian activities. Finally, it is necessary to test the algorithms in moving vehicles and cluttered backgrounds.

## Acknowledgment

This work was funded by Research Grants SEGVAUTO S2013/MIT-2713 (CAM), DPI2014-59276-R (Spanish Min. of Economy), BRAVE Project, H2020, Contract #723021. This project has received funding from the Electronic Component Systems for European Leadership Joint Undertaking under grant agreement No 737469. This Joint Undertaking receives support from the European Union Horizon 2020 research and innovation programme and Germany, Austria, Spain, Italy, Latvia, Belgium, Netherlands, Sweden, Finland, Lithuania, Czech Republic, Romania, Norway.